\documentclass[runningheads]{llncs}
\usepackage{graphicx}
\title{Conditional Inference and Activation of Knowledge Entities in ACT-R}

\author{%
Marco Wilhelm\inst{1}\orcidID{0000-0003-0266-2334} \and
Diana Howey\inst{1}\orcidID{0000-0002-7203-4862} \and
Gabriele Kern-Isberner\inst{1}\orcidID{0000-0001-8689-5391} \and
Kai Sauerwald\inst{2}\orcidID{0000-0002-1551-7016} \and
Christoph Beierle\inst{2}\orcidID{0000-0002-0736-8516}}

\authorrunning{M. Wilhelm, D. Howey, G. Kern-Isberner, K. Sauerwald, and C. Beierle}

\institute{%
Dept. of Computer Science, TU Dortmund University, 
Dortmund, Germany\\
\email{\{marco.wilhelm, diana.howey, 
gabriele.kern-isberner\}@cs.tu-dortmund.de} \and
Dept. of Computer Science, FernUniversität in Hagen,
Hagen, Germany\\
\email{\{kai.sauerwald, christoph.beierle\}@fernuni-hagen.de\}}}
\usepackage{tikz}
\usepackage{tabularx,booktabs}
\newcolumntype{Y}{>{\centering\arraybackslash}X}
\newcolumntype{R}{>{\raggedleft\arraybackslash}X}
\usepackage{amsmath,amssymb,bm}
\usepackage{nicefrac}

\usepackage{pgffor}

\foreach \x in {A,...,Z}{%
\expandafter\xdef\csname cal\x\endcsname{\noexpand\ensuremath{\noexpand\mathcal{\x}}}
\expandafter\xdef\csname frak\x\endcsname{\noexpand\ensuremath{\noexpand\mathfrak{\x}}}
\expandafter\xdef\csname bb\x\endcsname{\noexpand\ensuremath{\noexpand\mathbb{\x}}}
\expandafter\xdef\csname sf\x\endcsname{\noexpand\ensuremath{\noexpand\mathsf{\x}}}
}

\foreach \x in {a,...,z}{%
\expandafter\xdef\csname cal\x\endcsname{\noexpand\ensuremath{\noexpand\mathcal{\x}}}
\expandafter\xdef\csname frak\x\endcsname{\noexpand\ensuremath{\noexpand\mathfrak{\x}}}
\expandafter\xdef\csname bb\x\endcsname{\noexpand\ensuremath{\noexpand\mathbb{\x}}}
\expandafter\xdef\csname sf\x\endcsname{\noexpand\ensuremath{\noexpand\mathsf{\x}}}
}

\foreach \x in {act,amphibian,Bel,bird,egg,ex,new,oviparous,penguin,SAN,SAT}{%
\expandafter\xdef\csname \x\endcsname{\noexpand\ensuremath{\noexpand\mathsf{\x}}}
}

\foreach \x in {}{%
\expandafter\xdef\csname \x\endcsname{\noexpand\textbf{\x}}
}

\foreach \x in {true,false}{%
\expandafter\xdef\csname \x\endcsname{\noexpand\texttt{\x}}
}

\newcommand{\nminf}[2][]{\ensuremath{{\ |\hspace{-.5em}\sim^{#1}_{#2}\, }}}

\newcommand{\N}[1]{\mkern1.5mu\overline{\mkern-1.5mu#1\mkern-1.5mu}\mkern 1.5mu}
\newcommand{\nomath}[1]{~~~\text{#1}~~~ }
\renewcommand{\implies}{\Rightarrow}

\newcommand{\IFF}{\nomath{iff}}
\newcommand{\OTHERWISE}{\nomath{otherwise}}

\begin{document}
\maketitle

\begin{abstract}
Activation-based conditional inference
applies conditional reasoning 
to ACT-R, 
a cognitive architecture developed 
to formalize human reasoning.
The idea of activation-based conditional inference 
is to determine a reasonable
subset of a conditional belief base  
in order to draw inductive inferences in time. 
Central to activation-based conditional inference 
is the activation function which assigns to the 
conditionals in the belief base a degree of activation 
mainly based on the conditional's relevance 
for the current query and its usage history. 
Therewith, our approach integrates 
several aspects of human reasoning 
into expert systems 
such as focusing, forgetting, and remembering.
\end{abstract}
\keywords{Conditional Reasoning, Inductive Inference, Cognitive Model,\\
ACT-R, Focusing, Forgetting, Remembering}

\section{Introduction}
\emph{Expert systems} 
\cite{Jackson_1999,Rajendra_Sajja_2009}
are computer programs 
which infer implicit information from belief bases 
in order to solve complex reasoning tasks. 
Basically, expert systems
consist of two components, 
a belief base and an inference engine. 
When the user inserts a query, 
the belief base and the query 
are transferred to the inference engine 
which decides whether the query can be 
drawn from the belief base or not.
The aspiration of expert systems 
is to draw inferences of high quality 
from usually incomplete and uncertain beliefs.
 
The contribution of \emph{activation-based 
conditional inference}
to this inference process 
is a preselection of
beliefs, yielding a reduced belief base that
is transferred to the inference engine. 
The objective is both to  
reduce computational costs during the inference
process, and to
model human cognitive processes 
with expert systems more adequately.
It is obvious and reasonable that human reasoners
do not draw inferences based 
on all of their beliefs, in particular when they have to
make snap decisions in time. Basically, there are two
cognitive processes which affect the selection of 
beliefs: The long-term 
process
of forgetting and
remembering 
and the short-term process of activating certain beliefs
depending on the context.
In \emph{ACT-R}
(\emph{Adaptive Control of Thought-Rational}, 
\cite{Anderson_Lebiere_1998,Anderson_2007}), 
a
well-founded cognitive
architecture established in cognitive science in order
to formalize human reasoning,
the long-term memory is
represented by the \emph{base-level
activation},
while the 
context-dependent activation of beliefs is described
by
the 
\emph{spreading activation theory} \cite{Anderson_1983}.
The core idea behind the spreading activation theory
is that an initial priming caused
by sensory stimuli triggers certain 
\emph{cognitive units} \cite{Anderson_1983}
which again
trigger related cognitive units and so on until
the disposition for activation is too low to trigger
further cognitive units. 
The triggered cognitive units 
settle the current focus in which
reasoning takes place. 

In this paper, we adapt the concept of (de)activation
of knowledge entities 
from ACT-R and combine it with
conditional inference formalisms 
from nonmonotonic reasoning. 
More precisely,
we define a model for \emph{activation-based
inference} 
from conditional belief bases by 
adapting the
activation function from ACT-R to
conditionals 
of the form $(B|A)$ with the meaning
``if $A$ holds, then usually $B$ holds, too."
Therewith, on the one hand, we
generalize the concept of \emph{focused inference}
\cite{Wilhelm_Kern-Isberner_2021} and give it a
profound cognitive meaning.
And on the other hand, we equip ACT-R, which is typically realized as a \emph{production system}~\cite{Klahr_Langley_Neches_1987,Newell_1990}, with a modern inference formalism of high quality.

The rest of the paper is organized as follows. First,
we recall some basics on conditional logics, inductive
inference formalisms, and the ACT-R architecture. Then,
we give a brief outline of activation-based conditional
inference which is based on the activation function for conditionals. This activation function is examined in more
detail afterwards. Finally, we show how the concepts of
forgetting and remembering can be integrated into our framework before we conclude with an outlook.

\section{Preliminaries}
\subsection{Logical Foundations}
We consider a \emph{propositional language} $\calL$
which is defined 
over a finite set of \emph{propositional variables} (or \emph{atoms})~$\Sigma$.
Formulas in $\calL$ are built by the
common connectives $\neg$ (\emph{negation}), $\land$ (\emph{conjunction}), and $\lor$ (\emph{disjunction}).
The semantics of formulas in $\calL$ is given by 
\emph{interpretations} $I\in\calI$ as usual. 
We further 
use the abbreviations $AB=A\land B$, $\N{A}=\neg A$, 
$A\implies B=\N{A}\lor B$, $\top=A\lor\N{A}$, 
and $\bot=A\N{A}$.
An expression of the form $(B|A)$ with $A,B\in\calL$ is
called \emph{conditional} and has the intuitive meaning
``if $A$ holds, then usually $B$ holds, too."
Formally, conditionals are interpreted by
\emph{ranking functions} over \emph{possible worlds}
\cite{Spohn_2012}.
Here, possible worlds are the interpretations in $\calI$ 
represented as complete conjunctions of \emph{literals},
i.e. an atom or its negation. 
The set of all possible worlds is denoted 
by~$\Omega$.
A \emph{ranking function} $\kappa:\Omega\to\bbN_0^\infty$
maps possible worlds to a degree of plausibility 
while satisfying the normalization condition 
$\kappa^{-1}(0)\neq\emptyset$. 
Lower ranks indicate higher plausibility so that $\kappa^{-1}(0)$ is the set of the most plausible worlds.
The \emph{rank} of a formula $A$ is the minimal rank
of its models,
\mbox{
$\kappa(A)=\min\{\kappa(\omega)\mid\omega\in\Omega,\ \omega\models A\}$,}
where the convention $\min\emptyset=\infty$ applies.
$\kappa$~\emph{accepts} a conditional $(B|A)$
iff~$\kappa(AB)<\kappa(A\N{B})$
or $\kappa(A)=\infty$. 
A ranking function $\kappa$ is a \emph{model} of 
a \emph{belief base} $\Delta$, i.e. a finite set of conditionals, 
iff~$\kappa$
accepts all conditionals in~$\Delta$. 
A belief base is called \emph{consistent} iff
it has at least one model. 
The set of all belief bases over $\calL$ is denoted by~$\calD$.
If $X$ is a formula, a conditional,
or a belief base, then we denote the set of atoms mentioned
in $X$ with $\Sigma(X)$, i.e., $\Sigma(X)$ is the
\emph{signature} of~$X$.

\begin{table}[t]
\setlength{\tabcolsep}{12pt}
\begin{tabularx}{\columnwidth}{lX}
\toprule
Conditional & Meaning \\
\midrule
$\frakr_{1\hphantom{1}}=(f|aw)$
& Winged animals usually fly. \\
$\frakr_{2\hphantom{1}}=(\N{f}|a\N{w})$
& Wingless animals usually do not fly. \\
$\frakr_{3\hphantom{1}}=(b\implies a|\top)$
& Birds are animals. \\
$\frakr_{4\hphantom{1}}=(w|b)$
& Birds usually have wings. \\
$\frakr_{5\hphantom{1}}=(d|b)$
& Birds usually drink water. \\
\midrule
$\frakr_{6\hphantom{1}}=(p\implies b|\top)$
& Penguins are birds. \\
$\frakr_{7\hphantom{1}}=(\N{f}|p)$
& Penguins usually do not fly. \\
$\frakr_{8\hphantom{1}}=(c\implies b|\top)$
& Chickens are birds. \\
$\frakr_{9\hphantom{1}}=(\N{f}|c)$
& Chickens usually do not fly. \\
$\frakr_{10}=(f|cs)$
& Scared chickens usually fly. \\
\midrule
$\frakr_{11}=(\N{s}|c)$
& Chickens are usually not scared. \\
$\frakr_{12}=(i\implies a|\top)$
& Fish are animals. \\
$\frakr_{13}=(r\implies i|\top)$
& Freshwater fish are fish. \\
$\frakr_{14}=(l\implies i|\top)$
& Saltwater fish are fish. \\
$\frakr_{15}=(l\lor r|i)$
& Fish are usually saltwater fish or freshwater fish. \\
\midrule
$\frakr_{16}=(\N{d}|r)$
& Freshwater fish usually do not drink water. \\
$\frakr_{17}=(d|l)$
& Saltwater fish usually drink water. \\
$\frakr_{18}=(h\implies r|\top)$
& Hatchetfish are freshwater fish. \\
$\frakr_{19}=(f\N{w}|h)$
& Hatchetfish usually fly but are wingless. \\
$\frakr_{20}=(k\implies m|\top)$
& Kangaroos are marsupials. \\
\bottomrule
\end{tabularx}
\caption{Belief base $\Delta'=\{\frakr_1,\ldots,\frakr_{20}\}$ (cf. Example \ref{ex:example_knowledge_base}).}
\label{tab:example_knowledge_base}
\end{table}

\begin{example}
\label{ex:example_knowledge_base}
In 
Table~\ref{tab:example_knowledge_base}
an example of a consistent belief base over the
signature
$\Sigma'=\{a,b,c,d,f,h,i,k,l,m,p,r,s,w\}$ is shown.
For example, $\Sigma(\Delta)=\Sigma'$ and 
$\Sigma(\frakr_{6})=\{b,p\}$.
\end{example}

\subsection{Conditional Inference}
We now consider the task of drawing 
\emph{inductive inferences} from a belief base
$\Delta$ and recall
the notion of \emph{focused inference} from
\cite{Wilhelm_Kern-Isberner_2021}.
Roughly said, inductive inferences are
conditionals which are plausible consequences from $\Delta$.

\begin{definition}
(cf. \cite{Kern-Isberner_Beierle_Brewka_2020})
An \emph{(inductive) inference operator} $\frakI:\calD\to
\calL\times\calL$
is a mapping
which assigns
to each belief base $\Delta\in\calD$
an inference relation $\nminf[\frakI]{\Delta}\subseteq \calL\times\calL$
such that:
\begin{itemize}
\item If $(B|A)\in\Delta$, then $A\nminf[\frakI]{\Delta} B$.
\hfill (Direct Inference)
\item If $\Delta=\emptyset$, then $A\nminf[\frakI]{\Delta} B$
only if $A\models B$.
\hfill (Trivial Vacuity)
\end{itemize}
$\frakI_\Delta=\{(B|A)\mid A\nminf[\frakI]{\Delta} B\}$ denotes the
set of \emph{inductive inferences} from $\Delta$ wrt. $\frakI$. We further define
a three-valued \emph{inference response} to a \emph{query conditional}~$(B|A)$ by
\begin{equation*}
[\![(B|A)]\!]^\frakI_\Delta=
\begin{cases}
\text{yes}		& \IFF (B|A)\in\frakI_\Delta \\
\text{no}		& \IFF (\N{B}|A)\in\frakI_\Delta \\
\text{unknown}	& \OTHERWISE
\end{cases}.
\end{equation*}
\end{definition}

An important representative of
inference operators is the
\emph{System~P inference operator} $\frakI^P$, which
is defined by $(B|A)\in\frakI^P_\Delta$
iff every model of $\Delta$ accepts~$(B|A)$.
$\frakI^P$ is characterized by
a collection of inference rules 
which are well-established in nonmonotonic reasoning
\cite{Adams_1975,Kraus_Lehmann_Magidor_1990}. One also has \mbox{$(B|A)\in\frakI^P_\Delta$}
iff $\Delta\cup\{(\N{B}|A)\}$ is inconsistent
\cite{Goldszmidt_Pearl_1991}. 

Another well-founded inference operator 
is provided by \emph{System~Z} \cite{Pearl_1990}. Here, we are not interested in the inference operator $\frakI^Z$ itself but in the \emph{Z-partition} of $\Delta$ which is an
auxiliary structure for computing $\frakI^Z_\Delta$.
Z-partitions are ordered partitions of belief bases based on the notion of \emph{tolerance}.
A belief base $\Delta$ \emph{tolerates}
a conditional $(B|A)$ iff there is a possible world $\omega$
in which $(B|A)$ is \emph{verified}, i.e. $\omega\models AB$, and no conditional
from $\Delta$ is \emph{falsified}, i.e. $\omega\models (A'\implies B')$ for all $(B'|A')\in\Delta$.
An ordered partition
$(\Delta_0,\Delta_1,\ldots,\Delta_m)$ of $\Delta$
is a \emph{tolerance partition} of $\Delta$ iff
for $i=0,\ldots,m$ every conditional in $\Delta_i$ is tolerated by $\bigcup_{j=i}^m \Delta_j$.
The \emph{Z-partition} $Z(\Delta)$ is the unique tolerance
partition of $\Delta$ which is obtained by iteratively 
determining~$\Delta_i$ as the maximal set of tolerated conditionals.
If a conditional~$\frakr$ is in the $i$-th partition of $Z(\Delta)$, we say that~$\frakr$ has \emph{Z-rank} $Z_\Delta(\frakr)=i$. System Z
satisfies the paradigm of \emph{maximum normality}, i.e., the lower the Z-rank of a conditional is, the more normal the conditional is.

\begin{example}
\label{ex:normality}
The Z-ranks of the conditionals in $\Delta'$ 
(Table~\ref{tab:example_knowledge_base}) are shown in Table~\ref{tab:example_knowledge_base_values}. For example,
the Z-rank of $\frakr_1$ is $Z_{\Delta'}(\frakr_1)=0$ because
$\frakr_1$ is tolerated by $\Delta'$ (consider  $\omega=ab\N{c}df\N{h}\N{i}\N{k}\N{l}\N{m}\N{p}\N{r}\N{s}w$). Further, the Z-ranks
$Z_{\Delta'}(\frakr_{10})=2$ and
$Z_{\Delta'}(\frakr_9)=1$ illustrate the concept of normality. While conditional $\frakr_9$ is concerned about the
flight ability of chickens in general, conditional $\frakr_{10}$ makes a statement about the flight behavior of chicken when
they are in a special mood. Hence, conditional $\frakr_{10}$ applies to a more specific case than $\frakr_9$ 
and accordingly has a higher Z-rank.
\end{example}

An inference operator $\frakI$ is 
\emph{semi-monotonous}
iff for every two belief
bases $\Delta$ and $\tilde{\Delta}$ it holds that 
$\tilde{\Delta}\subseteq\Delta$ implies  
\mbox{$\frakI_{\tilde{\Delta}}\subseteq\frakI_{\Delta}$.}
While System~P inference is semi-monotonous
(cf., e.g., \cite{Wilhelm_Kern-Isberner_2021}),
System Z inference is not. We give an example
which illustrates the semi-monotony of System P.

\begin{example}
\label{ex:system_p}
Consider 
$\Delta''=\{\frakr_9,\frakr_{11}\}\subseteq\Delta'$ 
(Table \ref{tab:example_knowledge_base}).
Since $\Delta''\cup\{(f|c\N{s})\}$ is inconsistent,
$c\N{s} \nminf[P]{\Delta''} \N{f}$ follows. 
That is, one can
infer from $\Delta''$ wrt. System P 
that chicken which are not scared usually do not fly. 
Due to the semi-monotony of System P, 
this inference can also be drawn from 
$\Delta'$ because of $\Delta''\subseteq\Delta'$.
\end{example}

We now recall the concept of 
\emph{focused inference} from
\cite{Wilhelm_Kern-Isberner_2021}. The idea behind focused
inference is to draw inferences from a reasonable (as small as possible) subset of~$\Delta$ in order to make snap
but still well-founded decisions in time.
In this context,
the advantage of semi-monotonous 
inference operators like $\calI^P$
is that one does not risk to draw
false inferences when focusing on a subset $\tilde{\Delta}\subset\Delta$ because $[\![(B|A)]\!]^{\calI^P}_{\tilde{\Delta}}=\ \textit{yes (resp. no)}$ implies $[\![(B|A)]\!]^{\calI^P}_{\Delta}=\ \textit{yes (resp. no)}$.
In order to formalize focused inference, we consider mappings $\phi:\calD\to\calD$ with $\phi(\Delta)\subseteq\Delta$, i.e. mappings which return
subsets of $\Delta$.
We call such a mapping $\phi$ a \emph{focus}.

\begin{definition}
Let $\Delta$ be a belief base, 
$(B|A)$ a conditional,
$\frakI$ an inference operator,
and $\phi$ a focus. Then,  
\emph{$(B|A)$
\emph{follows} from $\Delta$
wrt.~$\frakI$
in the focus $\phi$} iff $(B|A)\in\frakI_{\phi(\Delta)}$.
\end{definition}

In \cite{Wilhelm_Kern-Isberner_2021}, the focus 
$\phi$ is defined iteratively based on the query $\frakq=(B|A)$:
The 
conditionals in the 
\emph{direct focus} $\phi^\frakq_0$ are those conditionals which share 
at least one atom with 
$\frakq$, i.e.
$\phi^\frakq_0(\Delta)=\{\frakr\in\Delta\mid 
\Sigma(\frakr)\cap\Sigma(\frakq)\neq\emptyset\}$.
The conditionals 
in the \emph{$i$-th focus} are determined by
$\phi^\frakq_i(\Delta)=\{\frakr\in\Delta\mid \exists \frakr\in\phi^\frakq_{i-1}(\Delta): \Sigma(\frakr)\cap\Sigma(\frakq)\neq\emptyset\}$.

\begin{example}
\label{ex:focused_inference}
The direct focus
of $\Delta'$ 
(Table \ref{tab:example_knowledge_base})
 wrt. $\frakq=(f|c\N{s})$  is
$\Delta_0=\phi^{\frakq}_0(\Delta')$ with
$\Delta_0=\{\frakr_{1},\frakr_{2},\frakr_{7},\frakr_{8},\frakr_{9},\frakr_{10},\frakr_{11},\frakr_{19}\}$.
One has $[\![\frakq]\!]^{\frakI^P}_{\Delta_0}=\textit{no}$.
According to Example~\ref{ex:system_p},
one already has $[\![\frakq]\!]^{\frakI^P}_{\Delta''}=\textit{no}$
where 
$\Delta''=\{\frakr_9,\frakr_{11}\}\subset\Delta_0$, though.
Hence, the direct focus does not have to be the smallest possible focus in which an inference can be drawn.
On the contrary, a focus can also be too small in order to decide a query. For instance, 
$[\![\frakq]\!]^{\frakI^P}_{\Delta'''}=\textit{unknown}$ wrt. any $\Delta'''\subset\Delta'$ with $\{\frakr_9,\frakr_{11}\}\not\subseteq\Delta'''$.
\end{example}

Apart from
the computational benefits of drawing inferences wrt. small foci, appropriate foci are
also interesting from the knowledge representation and reasoning (KRR) perspective because they unveil the part of the belief base which is relevant for answering the query.
Unfortunately, finding appropriate foci is challenging.
In the following, we approach this problem from the
cognitive science perspective and develop a framework for 
drawing focused inferences which are justified by cognitive
principles.

\subsection{ACT-R Architecture}
\emph{ACT-R}
\cite{Anderson_Lebiere_1998,Anderson_2007}
is a 
production systems based 
cognitive architecture which
formalizes human reasoning.
In ACT-R a distinction is made 
between \emph{declarative} and \emph{procedural} memory. 
In the declarative memory, categorical knowledge about 
individuals or objects is stored in form of \emph{chunks}
(\emph{knowing that}) while the procedural memory 
consists of \emph{production rules}
and describes how chunks are processed 
(\emph{knowing how}, \cite{Ryle_2000}).
Reasoning in ACT-R starts with an 
initial \emph{priming}, for example
a stimulus from the environment, which
causes an activation of chunks. The chunk with the highest activation is processed by production rules 
in order to compute a solution to the reasoning task.
If this fails, 
the activation passes into
an iterative process:
The system obtains additional chunks from the declarative 
memory and tries to compute a solution again.
The iteration stops when either the problem is solved
or no further chunks are active.
The retrieval of chunks is a very 
refined process in \mbox{ACT-R}.
Basically, it depends on an 
\emph{activation function} 
which is calculated for each specific request
and is based on a \emph{usage history} of the chunks,
associations between \emph{cognitive units}
and the \emph{priming}~\cite{Anderson_1983}.
There is no clear consensus about the kind of cognitive units despite of the perception
that they form the basic building blocks of thinking \cite{Anderson_1980}.

How the activation of a chunk $\calA(\frakc_i)$
is computed in detail 
depends on multiple parameters and the configurations 
of the ACT-R system, but is mainly given by the 
sum of the so-called \emph{base-level activation} 
$\calB(\frakc_i)$
and the \emph{spreading activation}~$\calS(\frakc_i)$,
which again is a sum
of \emph{degrees of associations} between chunks~
$\calS(\frakc_i,\frakc_j)$ weighted by some
\emph{weighting factors} $\calW(\frakc_j)$:
\begin{equation}
\label{eq:activation_function}
\calA(\frakc_i)=\calB(\frakc_i)+
\sum_{j} \calW(\frakc_j)\cdot \calS(\frakc_i,\frakc_j).
\end{equation}
%
The \emph{base-level activation} of a chunk $\calB(\frakc_i)$
reflects the \emph{entrenchment} of $\frakc_i$ in the reasoner's memory and
depends on the recency and frequency of its use. Typically, $\calB(\frakc_i)$
is 
decreased
over time (\emph{fading out}) 
and is increased when the chunk is active.
Further, $\calB(\frakc_i)$ is independent of
the priming.

In contrast,
the \emph{spreading activation} of a chunk $\calS(\frakc_i)$ depends on the priming and exploits the
well-known \emph{spreading activation theory}~\cite{Anderson_1983} to formalize
how the brain iterates through a network 
of associated ideas to retrieve information.
In the spreading activation theory one breaks 
down the notion of ideas into \emph{cognitive units}. 
Usually, the cognitive units
are arranged as vertices in an undirected graph,
the so-called \emph{spreading activation network}~$\calN(\Delta)$, and an initial \emph{triggering} of some cognitive units caused by the priming is propagated through~$\calN(\Delta)$. The spreading activation $\calS(\frakc_i)$ can then be derived from
the \emph{triggering values} of the cognitive units of which $\frakc_i$ makes use. 
The interrelation of cognitive units and of chunks is specified in more detail in the \emph{degree of association}
and the \emph{weighting factor}.

The \emph{degree of association} $\calS(\frakc_i,\frakc_j)$ reflects
how strongly related~$\frakc_i$ and $\frakc_j$ are. 
Chunks which deal with the same issue have a 
high degree of
association while chunks which refer to different
topics are only loosely or not related and, therefore, have a low degree of association.
Technically,~$\calS(\frakc_i,\frakc_j)$ is based on the cognitive units which $\frakc_i$ and~$\frakc_j$ have in common.
The degrees of association are weighted by 
the \emph{weighting factors} $\calW(\frakc_i)$. 
While the degree of association 
is independent of the priming,
the weighting factors reflect the context-dependency
of $\calA(\frakc_i)$.
Only if $\frakc_i$ is associated to a
chunk $\frakc_j$ ($\calS(\frakc_i,\frakc_j)>0$) 
which 
has positive weight ($\calW(\frakc_j)>0$),
then the chunk $\frakc_i$ has a positive
spreading activation~($\calS(\frakc_i)>0)$, too.

\section{Activation-Based Conditional Inference}

\begin{table*}[t]
\setlength{\tabcolsep}{3pt}
\begin{tabularx}{\textwidth}{l|YY|YYY|YYY}
\toprule
Conditional & $Z^{\Delta'}(\frakr)$ & $\calB^{\Delta'}(\frakr_i)$ & $\calW^{\Delta'}_{\frakq_1}(\frakr_i)$ &
$\calS^{\Delta'}_{\frakq_1}(\frakr_i)$ &
$\calA^{\Delta'}_{\frakq_1}(\frakr_i)$
& $\calW^{\Delta'}_{\frakq_2}(\frakr_i)$ &
$\calS^{\Delta'}_{\frakq_2}(\frakr_i)$ &
$\calA^{\Delta'}_{\frakq_2}(\frakr_i)$
\\
\midrule
$\frakr_1$ 				
& $0$ & $1$ & $\nicefrac{1}{3}$ & $1.36$
& $\boxed{2.36}$ & $\nicefrac{1}{4}$ & $1.27$ & $2.27$ \\
$\frakr_2$ 		
& $0$ & $1$ & $\nicefrac{1}{3}$ & $1.36$
& $\boxed{2.36}$ & $\nicefrac{1}{4}$ & $1.27$ & $2.27$ \\
$\frakr_3$ 	
& $0$ & $1$ & $\nicefrac{2}{3}$ & $1.41$
& $\boxed{2.41}$ & $\nicefrac{1}{4}$ & $0.66$ & $1.66$ \\
$\frakr_4$ 				
& $0$ & $1$ & $\nicefrac{1}{3}$ & $1.13$ 
& $2.13$ & $\nicefrac{1}{4}$ & $0.70$ & $1.70$ \\
$\frakr_5$ 				
& $0$ & $1$ & $\nicefrac{2}{15}$ & $0.82$ 
& $1.82$ & $\nicefrac{1}{21}$ & $0.41$ & $1.41$ \\
\midrule
$\frakr_6$ 	
& $0$ & $1$ & $\nicefrac{2}{3}$ & $1.36$
& $\boxed{2.36}$ & $\nicefrac{1}{4}$ & $0.60$ & $1.60$ \\
$\frakr_7$ 			
& $1$ & $\nicefrac{1}{2}$ & $\nicefrac{2}{3}$ & $1.23$ 
& $1.73$ & $\nicefrac{1}{4}$ & $1.10$ & $1.60$ \\
$\frakr_8$ 	
& $0$ & $1$ & $\nicefrac{4}{15}$ & $1.03$
& $2.03$ & $\nicefrac{1}{4}$ & $1.43$ & $\boxed{2.43}$ \\
$\frakr_9$ 			
& $1$ & $\nicefrac{1}{2}$ & $\nicefrac{4}{15}$ & $0.93$
& $1.43$ & $1$ & $2.35$ & $\boxed{2.85}$ \\
$\frakr_{10}$ 			
& $2$ & $\nicefrac{1}{3}$ & $\nicefrac{2}{15}$ & $0.81$
& $1.14$ & $1$ & $2.61$ & $\boxed{2.94}$ \\
\midrule
$\frakr_{11}$ 			
& $1$ & $\nicefrac{1}{2}$ & $\nicefrac{2}{15}$ & $0.40$
& $0.90$ & $1$ & $2.08$ & $\boxed{2.58}$ \\
$\frakr_{12}$ 
& $0$ & $1$ & $\nicefrac{1}{3}$ & $0.78$
& $1.78$ & $\nicefrac{1}{21}$ & $0.29$ & $1.29$ \\
$\frakr_{13}$ 
& $0$ & $1$ & $\nicefrac{1}{15}$ & $0.29$
& $1.29$ & $\nicefrac{1}{21}$ & $0.12$ & $1.12$ \\
$\frakr_{14}$ 
& $0$ & $1$ & $\nicefrac{1}{15}$ & $0.27$
& $1.27$ & $\nicefrac{4}{151}$ & $0.08$ & $1.08$ \\
$\frakr_{15}$ 		
& $0$ & $1$ & $\nicefrac{1}{15}$ & $0.29$
& $1.29$ & $\nicefrac{4}{151}$ & $0.12$ & $1.12$ \\
\midrule
$\frakr_{16}$ 			
& $0$ & $1$ & $\nicefrac{1}{15}$ & $0.19$
& $1.19$ & $\nicefrac{1}{21}$ & $0.11$ & $1.11$ \\
$\frakr_{17}$ 				
& $0$ & $1$ & $\nicefrac{1}{15}$ & $0.17$
& $1.17$ & $\nicefrac{4}{151}$ & $0.07$ & $1.07$ \\
$\frakr_{18}$ 
& $0$ & $1$ & $\nicefrac{1}{15}$ & $0.18$
& $1.18$ & $\nicefrac{1}{21}$ & $0.15$ & $1.15$ \\
$\frakr_{19}$ 		
& $1$ & $\nicefrac{1}{2}$ & $\nicefrac{1}{5}$ & $0.89$
& $1.39$ & $\nicefrac{1}{4}$ & $1.09$ & $1.59$ \\
$\frakr_{20}$ 
& $0$ & $1$ & $0$ & $0$ & $1$ & $0$ & $0$ & $1$ \\
\bottomrule
\end{tabularx}
\caption{Z-ranks $Z^{\Delta'}$, base-level activation
$\calB^{\Delta'}$, weighting factors $\calW^{\Delta'}_{\frakq_i}$,
spreading activation $\calS^{\Delta'}_{\frakq_i}$, and activation
function $\calA^{\Delta'}_{\frakq_i}$ wrt.
$\frakq_1=(p\implies a|\top)$ and $\frakq_2=(\N{f}| c\N{s})$
for the conditionals in $\Delta'$. Selected conditionals
are boxed (threshold $\theta=2.3$).}
\label{tab:example_knowledge_base_values}
\end{table*}

As common ACT-R implementations are
production systems which process chunks that are represented as simple lists of attributes, the logical basis of ACT-R does not hold the pace with modern KRR 
formalisms in nonmonotonic reasoning. Thus, we propose a 
cognitively inspired
model of inductive conditional reasoning
by interpreting the concepts
of ACT-R in terms of logic,
conditionals, and inference.
More precisely, we replace chunks by conditionals of a belief
base $\Delta$ and derive a focus $\phi$ based on the activation function in
(\ref{eq:activation_function}) in order to draw focused inferences wrt. an inference operator $\frakI_{\phi(\Delta)}$. 
Here, we rely on $\frakI^P_{\phi(\Delta)}$ because of the semi-monotony of System~P.
In our formalism,
atoms play the role of cognitive units, and the production rules are replaced by the inference operator. 
From the conditional logical perspective,
the added value of this \emph{activation-based conditional 
inference} approach are 
\begin{itemize}
\item the
cognitive justification of the focus, 
\item the possibility
of a more fine-grained adjustment of the focus than in
\cite{Wilhelm_Kern-Isberner_2021},
\item and the option to 
integrate further cognitive concepts such as 
forgetting and remembering.
\end{itemize}
Formally,
we calculate an activation value~$\calA(\frakr)>0$ for
every conditional $\frakr$ in $\Delta$.
If $\calA(\frakr)$ is above a certain
threshold~$\theta$, $\calA(\frakr)\geq\theta$, the conditional is selected
for the focus on $\phi(\Delta)$.
For this, we define a \emph{selection function}
$s_\calA^\theta:\Delta\to\{0,1\}$ with
$s_\calA^\theta(\frakr)=1$ iff $\calA(\frakr)\geq \theta$
and $s_\calA^\theta(\frakr)=0$ otherwise. We
denote the set of selected conditionals
by 
\[
\Delta_\calA^\theta=\{\frakr\in\Delta\mid
s_\calA^\theta(\frakr)=1\}.
\]
Note that $\Delta_\calA^\theta$ will
implicitly depend on
a query $\frakq=(B|A)$
since queries will serve as the initial priming
and the spreading activation, which is part of~$\calA$, depends on the priming.

\begin{definition}
Let~$\Delta$ be a belief base,
$(B|A)$ a conditional,
$\frakI$ an inference operator,
$\calA$ an activation function
for $\Delta$,
and $\theta\geq 0$. Then, $(B|A)$ is \emph{activation-based inferred} from $\Delta$ wrt. $\frakI$, $\calA$, and $\theta$ iff
$(B|A)\in\frakI_{\Delta^\theta_\calA}$.
\end{definition}

If answering
a query fails, i.e. $[\![(B|A)]\!]^\frakI_{\Delta^\theta_\calA}
=\textit{unknown}$, then the inference process can
be repeated by iteratively choosing a lower threshold 
$\theta_{i+1}<\theta_i$ which
leads to a larger (or equal) set of selected conditionals.
In the limit, when choosing $\theta=0$,
one has $\Delta^\theta_\calA=\Delta$, thus
\mbox{$\frakI_{\Delta^0_\calA}=\frakI_\Delta$}.
This iteration process is in analogy to
the sequence~$(\phi_i^\frakq(\Delta))_{i\in\bbN_0}$ defined
in \cite{Wilhelm_Kern-Isberner_2021}
and can be used to approximate~$\frakI_\Delta$
for any inductive inference operator $\frakI$.
In particular,
the chance of successfully answering
the query increases with each iteration step
when $\frakI$ is semi-monotonous.

\section{Blueprint for Activation Based Conditional Inference}

ACT-R does not formalize the activation function in (\ref{eq:activation_function}) in more detail but describes
its functionality informally. Hence, there is certain freedom in its configuration. We give a concrete instantiation of (\ref{eq:activation_function}) in the conditional inference setting which can be seen as a blue print for
further investigations and empirical analyses. 
Note that we shift the dependence of the base-level activation on the
usage history of conditionals to the next section.

Let $\Delta$
be a belief base, $\frakr_i\in\Delta$, and $\frakq$ a conditional (the query resp. priming). Then, (\ref{eq:activation_function}) becomes

\[
\calA_\frakq^\Delta(\frakr_i)=\underbrace{
\vphantom{\sum_{\frakr_j\in\Delta}}
\calB^\Delta(\frakr_i)}_{\text{base-level activation}}+\underbrace{\sum_{\frakr_j\in\Delta}
\calW_\frakq^\Delta(\frakr_j)
\cdot \calS(\frakr_i,\frakr_j)}_{\text{spreading
activation}\ \calS^{\Delta}_\frakq(\frakr_i)}.
\]

We explain the single components 
of $\calA^\Delta_\frakq(\frakr_i)$ in detail.

\subsection{Base-Level Activation}
$\calB^\Delta(\frakr)$ reflects
the entrenchment of $\frakr$ in the reasoner's
memory.
Since epistemic entrenchment and ranking semantics are dual
ratings, the \emph{normality} of a conditional is a good
estimator and we define
\[
\calB^\Delta(\frakr)=\frac{1}{1+Z^\Delta(\frakr)},
\qquad \frakr\in\Delta,
\]
where $Z^\Delta(\frakr)$ is the Z-rank of $\frakr$. 
Following this definition, 
$\calB^\Delta(\frakr)$ is positive and normalized by $1$.
While the most normal conditionals 
have a base-level activation
of $\calB^\Delta(\frakr)=1$, this value decreases with increasing 
specificity of $\frakr$.

\begin{example}
Table \ref{tab:example_knowledge_base_values} shows
the base-level activations of the conditionals in 
$\Delta'$ 
(Table~\ref{tab:example_knowledge_base}). 
For example, $\calB^{\Delta'}(\frakr_9)=\nicefrac{1}{2}$ and $\calB^{\Delta'}(\frakr_{10})=\nicefrac{1}{3}$.
Since $\frakr_9$ is less specific than 
$\frakr_{10}$ (cf. Example \ref{ex:normality}), its base-level activation is higher than
$\calB^{\Delta'}(\frakr_{10})$.
\end{example}

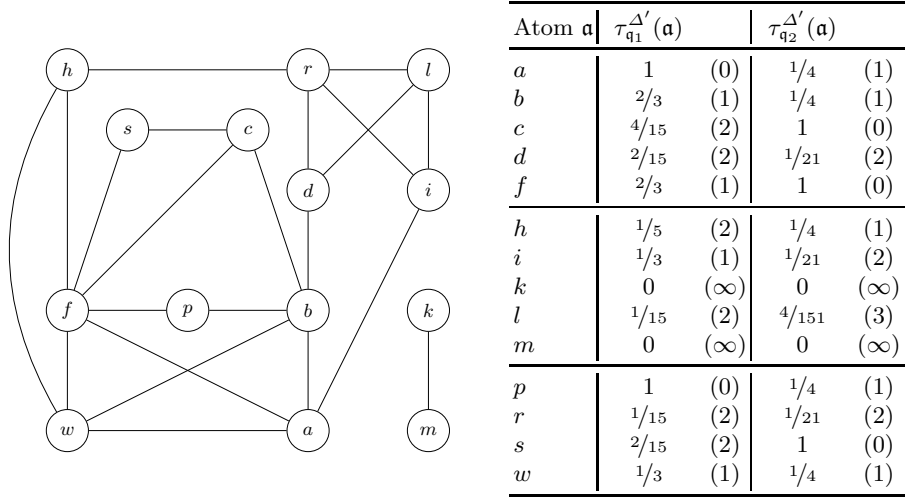
\begin{figure*}[t]
\begin{minipage}{.45\textwidth}
\begin{center}
\scalebox{.8}{
\begin{tikzpicture}[node distance=2cm,every node/.style={draw,circle,minimum size=.7cm}]
\node (a) {$a$};
\node[above of=a] (b) {$b$};
\node[left of=b] (p) {$p$};
\node[above of=p,xshift=1cm,yshift=1cm] (c) {$c$};
\node[left of=p] (f) {$f$};
\node[above of=b] (d) {$d$};
\node[right of=d] (i) {$i$};
\node[above of=d] (r) {$r$};
\node[right of=b] (k) {$k$};
\node[right of=r] (l) {$l$};
\node[below of=k] (m) {$m$};
\node[left of=c] (s) {$s$};
\node[below of=f] (w) {$w$};
\node[above of=s,xshift=-1cm,yshift=-1cm] (h) {$h$};
\draw (a) -- (b);
\draw (a) -- (f);
\draw (a) -- (i);
\draw (a) -- (w);
\draw (b) -- (c);
\draw (b) -- (d);
\draw (b) -- (p);
\draw (b) -- (w);
\draw (c) -- (f);
\draw (c) -- (s);
\draw (d) -- (l);
\draw (d) -- (r);
\draw (f) -- (h);
\draw (f) -- (p);
\draw (f) -- (s);
\draw (f) -- (w);
\draw (h) -- (r);
\draw (h) to[bend right] (w);
\draw (i) -- (l);
\draw (i) -- (r);
\draw (k) -- (m);
\draw (l) -- (r);
\end{tikzpicture}
}
\end{center}
\end{minipage}
\hfill
\begin{minipage}{.43\textwidth}
\begin{tabularx}{\textwidth}{l|Yc|Yc}
\toprule
Atom $\fraka$ & $\tau^{\Delta'}_{\frakq_1}(\fraka)$ & & 
$\tau^{\Delta'}_{\frakq_2}(\fraka)$ & \\
\midrule
$a$ & $1$ & $(0)$ &
$\nicefrac{1}{4}$ & $(1)$ \\
$b$ & $\nicefrac{2}{3}$ & $(1)$ &
$\nicefrac{1}{4}$ & $(1)$ \\
$c$ & $\nicefrac{4}{15}$ & $(2)$ &
$1$ & $(0)$ \\
$d$ & $\nicefrac{2}{15}$ & $(2)$ &
$\nicefrac{1}{21}$ & $(2)$ \\
$f$ & $\nicefrac{2}{3}$ & $(1)$ &
$1$ & $(0)$ \\
\midrule
$h$ & $\nicefrac{1}{5}$ & $(2)$ &
$\nicefrac{1}{4}$ & $(1)$ \\
$i$ & $\nicefrac{1}{3}$ & $(1)$ &
$\nicefrac{1}{21}$ & $(2)$ \\
$k$ & $0$ & $(\infty)$ &
$0$ & $(\infty)$ \\
$l$ & $\nicefrac{1}{15}$ & $(2)$ &
$\nicefrac{4}{151}$ & $(3)$ \\
$m$ & $0$ & $(\infty)$ &
$0$ & $(\infty)$ \\
\midrule
$p$ & $1$ & $(0)$ &
$\nicefrac{1}{4}$ & $(1)$ \\
$r$ & $\nicefrac{1}{15}$ & $(2)$ &
$\nicefrac{1}{21}$ & $(2)$ \\
$s$ & $\nicefrac{2}{15}$ & $(2)$ &
$1$ & $(0)$ \\
$w$ & $\nicefrac{1}{3}$ & $(1)$ &
$\nicefrac{1}{4}$ & $(1)$ \\
\bottomrule
\end{tabularx}
\end{minipage}
\caption{Unlabeled spreading activation network $\calN(\Delta')$ and labeling of $\calN(\Delta')$ wrt. the queries $\frakq_1=(p\implies a|\top)$ and $\frakq_2=(f|c\N{s})$. The numbers in the parentheses next to the labels (i.e., triggering values) are the
iteration steps in which the atoms are labeled. $0$ stands for the priming 
and $\infty$ for unreachable atoms. 
}
\label{fig:spreading_activation_network}
\end{figure*}

\subsection{Degree of Association}
$\calS(\frakr_i,\frakr_j)$ is a measure of connectedness 
between the conditionals in $\Delta$ and is defined by
\[
S(\frakr_i,\frakr_j)=
\frac{|\Sigma(\frakr_i)\cap\Sigma(\frakr_j)|}
{|\Sigma(\frakr_i)\cup\Sigma(\frakr_j)|},\qquad \frakr_i,\frakr_j\in\Delta.
\]
Hence, it is the number of shared atoms relative to all atoms in $\frakr_i$ or $\frakr_j$ 
and, therefore, non-negative and normalized by~$1$.
The degree of association
of a conditional $\frakr$ to itself is~$\calS(\frakr,\frakr)=1$ 
while the degree of
association of conditionals
which do not share any atoms is $0$.
The syntactically-driven definition of $\calS(\frakr_i,\frakr_j)$ is motivated by and extends the principle of 
\emph{relevance}
from nonmonotonic
reasoning. This principle of relevance
states that
if the belief base $\Delta$ splits into two sub-belief bases~$\Delta_1$ and $\Delta_2$ with 
$\Sigma(\Delta_1)\cap\Sigma(\Delta_2)=\emptyset$
and the query is defined over one of the 
signatures $\Sigma(\Delta_i)$, say $\Sigma(\Delta_1)$, only, then only 
the conditionals in $\Delta_1$
should be relevant for answering
this query \cite{Kern-Isberner_Beierle_Brewka_2020}.
Not only the quantities 
$\calS(\frakr_i,\frakr_j)$ for $\frakr_j\in\Delta$ themselves
are essential for the spreading activation of a 
conditional $\frakr_i$ 
but also 
how \emph{many} conditionals $\frakr_i$ 
is associated with.
The more a conditional is cross-linked
within $\Delta$,
the more likely it is that this conditional has a high
spreading activation 
and is selected by the $s$.

\begin{example}
The degrees of association between the conditionals in $\Delta'$ (Table~\ref{tab:example_knowledge_base}) are shown in Table \ref{tab:degree_of_association}.
For example,
\[
\calS(\frakr_{9},\frakr_{10})=\frac{|\{c,f\}\cap\{c,f,s\}|}{|\{c,f\}\cup\{c,f,s\}|}=\frac{|\{c,f\}|}{|\{c,f,s\}|}=\frac{2}{3}.
\]
\end{example}

\begin{table*}[h!]
\rotatebox{90}{
\renewcommand{\arraystretch}{1.3}
\begin{tabularx}{1.35\textwidth}{l|YYYYY|YYYYY|YYYYY|YYYYY}
\toprule
$\calS(\frakr_i,\frakr_j)$
& $\frakr_1$ & $\frakr_2$ & $\frakr_3$ & $\frakr_4$ & $\frakr_5$ & $\frakr_6$ & $\frakr_7$ & $\frakr_8$ & $\frakr_9$ & $\frakr_{10}$ & $\frakr_{11}$ & $\frakr_{12}$ & $\frakr_{13}$ & $\frakr_{14}$ & $\frakr_{15}$ & $\frakr_{16}$ & $\frakr_{17}$ & $\frakr_{18}$ & $\frakr_{19}$ & $\frakr_{20}$ \\
\midrule
$\frakr_{1}$
&$1$&$1$&$\nicefrac{1}{4}$&$\nicefrac{1}{4}$&&&$\nicefrac{1}{4}$&&$\nicefrac{1}{4}$&$\nicefrac{1}{5}$&&$\nicefrac{1}{4}$&&&&&&&$\nicefrac{1}{2}$& \\
$\frakr_{2}$
&&$1$&$\nicefrac{1}{4}$&$\nicefrac{1}{4}$&&&$\nicefrac{1}{4}$&&$\nicefrac{1}{4}$&$\nicefrac{1}{5}$&&$\nicefrac{1}{4}$&&&&&&&$\nicefrac{1}{2}$& \\
$\frakr_{3}$
&&&$1$&$\nicefrac{1}{3}$&$\nicefrac{1}{3}$&$\nicefrac{1}{3}$&&$\nicefrac{1}{3}$&&&&$\nicefrac{1}{3}$&&&&&&&& \\
$\frakr_{4}$
&&&&$1$&$\nicefrac{1}{3}$&$\nicefrac{1}{3}$&&$\nicefrac{1}{3}$&&&&&&&&&&&$\nicefrac{1}{4}$& \\
$\frakr_{5}$
&&&&&$1$&$\nicefrac{1}{3}$&&$\nicefrac{1}{3}$&&&&&&&&$\nicefrac{1}{3}$&$\nicefrac{1}{3}$&&& \\
\midrule
$\frakr_{6}$
&&&&&&$1$&$\nicefrac{1}{3}$&$\nicefrac{1}{3}$&&&&&&&&&&&& \\
$\frakr_{7}$
&&&&&&&$1$&&$\nicefrac{1}{3}$&$\nicefrac{1}{4}$&&&&&&&&&$\nicefrac{1}{4}$& \\
$\frakr_{8}$
&&&&&&&&$1$&$\nicefrac{1}{3}$&$\nicefrac{1}{4}$&$\nicefrac{1}{3}$&&&&&&&&& \\
$\frakr_{9}$
&&&&&&&&&$1$&$\nicefrac{2}{3}$&$\nicefrac{1}{3}$&&&&&&&&$\nicefrac{1}{4}$& \\
$\frakr_{10}$
&&&&&&&&&&$1$&$\nicefrac{2}{3}$&&&&&&&&$\nicefrac{1}{5}$& \\
\midrule
$\frakr_{11}$
&&&&&&&&&&&$1$&&&&&&&&& \\
$\frakr_{12}$
&&&&&&&&&&&&$1$&$\nicefrac{1}{3}$&$\nicefrac{1}{3}$&$\nicefrac{1}{4}$&&&&& \\
$\frakr_{13}$
&&&&&&&&&&&&&$1$&$\nicefrac{1}{3}$&$\nicefrac{2}{3}$&$\nicefrac{1}{3}$&&$\nicefrac{1}{3}$&& \\
$\frakr_{14}$
&&&&&&&&&&&&&&$1$&$\nicefrac{2}{3}$&&$\nicefrac{1}{3}$&&& \\
$\frakr_{15}$
&&&&&&&&&&&&&&&$1$&$\nicefrac{1}{4}$&$\nicefrac{1}{4}$&$\nicefrac{1}{4}$&& \\
\midrule
$\frakr_{16}$
&&&&&&&&&&&&&&&&$1$&$\nicefrac{1}{3}$&$\nicefrac{1}{3}$&& \\
$\frakr_{17}$
&&&&&&&&&&&&&&&&&$1$&&& \\
$\frakr_{18}$
&&&&&&&&&&&&&&&&&&$1$&$\nicefrac{1}{4}$& \\
$\frakr_{19}$
&&&&&&&&&&&&&&&&&&&$1$& \\
$\frakr_{20}$
&&&&&&&&&&&&&&&&&&&&$1$ \\
\bottomrule
\end{tabularx}
}
\caption{Degrees of association $\calS(\frakr_i,\frakr_j)$ between 
the conditionals $\frakr_i,\frakr_j\in\Delta'$. Since $\calS(\frakr_i,\frakr_j)$ is symmetric in its arguments, only the entries in the upper right triangle of the table are shown. Also $0$-entries are left out for a better readability.}
\label{tab:degree_of_association}
\end{table*}

\subsection{Weighting Factor}
$\calW^\Delta_\frakq(\frakr)$ indicates how
much the initial priming $\frakq$ 
triggers the conditional $\frakr$. We formalize the
influence of the priming according to the spreading activation theory by a labeling of the spreading activation network~$\calN(\Delta)$ between cognitive units.
In our context, the cognitive units are the atoms 
$\fraka\in\Sigma$ and the outcome of $\calN(\Delta)$ is a \emph{triggering value} $\tau^\Delta_\frakq(\fraka)\in[0,1]$ which indicates 
how much $\fraka$ is triggered by $\frakq$.
We follow the idea that a conditional $\frakr$ is
triggered not more than the atoms in $\Sigma(\frakr)$
and define the weighting factor by
\[
\calW^\Delta_\frakq(\frakr)=\min \{\tau^\Delta_\frakq(\fraka)
\mid \fraka\in\Sigma(\frakr)\}.
\]

\begin{example}
The weighting factors of the conditionals in $\Delta'$
(Table \ref{tab:example_knowledge_base}) wrt. queries $\frakq_1$ and $\frakq_2$ are shown in Table
\ref{tab:example_knowledge_base_values}. The
weighting factors depend on the labeling of the spreading activation network
in Figure \ref{fig:spreading_activation_network} which is explained in the next paragraph. 
For example, 
$\tau^{\Delta'}_{\frakq_1}(c)=\nicefrac{4}{15}$ and $\tau^{\Delta'}_{\frakq_1}(f)=\nicefrac{2}{3}$.
Consequently, the weighting factor of $\frakr_9$ wrt. $\frakq_1$ is
$\calW^{\Delta'}_{\frakq_1}(\frakr_9)=\min\{\nicefrac{4}{15},\nicefrac{2}{3}\}=\nicefrac{4}{15}$.
\end{example}

\subsection{Spreading Activation Network}
$\calN(\Delta)=(\calV,\calE)$ is an undirected graph 
with vertices~\mbox{$\calV=\Sigma$.}
Edges in $\calE$ represent
associations between the atoms in $\Sigma$ along which the
triggering of the atoms spreads. Two atoms are associated
if they occur commonly in some conditionals in $\Delta$, i.e.
\[
\calE=\{\{\fraka,\frakb\}\mid
\exists \frakr\in \Delta: \{\fraka,\frakb\}\subseteq\Sigma(\frakr)
\}.
\]
The actual spreading of activation is modeled by iteratively
labeling the vertices (atoms) in $\calN(\Delta)$
with their triggering value
$\tau^\Delta_\frakq(\fraka)$.
The \emph{labeling algorithm} is shown in 
Figure~\ref{fig:labeling_of_variables}. 
It starts with labeling
the atoms which 
are mentioned in the
query $\frakq$ with $1$. In the subsequent step, the
neighboring atoms are labeled and so on. 
The remaining atoms which are not reachable from the initially labeled atoms in $\Sigma(\frakq)$ are labeled with $0$. The labels of the atoms in between are the sum of the labels of the already labeled neighbors weighted by the sum of all labels so far plus $1$. This guarantees that these labels are between $0$ and $1$ and decrease for increasing iteration steps. Therewith, the triggering value of an atom depends on both the triggering values of the associated (sooner triggered) atoms and their count. 

\begin{figure}[t]
\begin{tabularx}{\columnwidth}{X}
\toprule
\textbf{Labeling Algorithm}\\
\midrule
\parbox{1.5cm}{\textbf{Input:}}  Spreading activation network $\calN(\Delta)=(\calV,\calE)$ (unlabeled);\\
\parbox{1.5cm}{\mbox{}} query $
\frakq=(B|A)$ \\
\parbox{1.5cm}{\textbf{Output:}} Labeling of $\calN(\Delta)$, i.e. triggering values $\tau^\Delta_\frakq(\fraka)=label(\fraka)$
for $\fraka\in\Sigma\ (=\calV)$ \\
\midrule
\parbox{.5cm}{1} \textbf{for} $\fraka\in\calV$ with $\fraka\in\Sigma(\frakq)$ \textbf{do}\\
\parbox{1cm}{2} $label(\fraka)=1$ \\
\parbox{.5cm}{3} \textbf{initialize} \\
\parbox{1cm}{4}
 $\calL\ =\{\fraka\in\calV\mid \fraka\ \text{is labeled}\}$, \\
\parbox{1cm}{5}
 $\calV'=\{\fraka\in\calV\mid \exists \{\fraka,\frakb\}\in\calE$: $\fraka\in\calV\setminus \calL
\land \frakb\in \calL \}$ \\
\parbox{.5cm}{6} \textbf{while} $\calV'\neq \emptyset$ \textbf{do} \\
\parbox{1cm}{7} \textbf{for} $\fraka\in\calV'$ \textbf{do} \\
\parbox{1.5cm}{8}
$
\displaystyle label(\fraka)=\frac{\sum_{\frakb\in \calL:\ \{\fraka,\frakb\}\in\calE}\ label(\frakb)}{1+\sum_{\frakb\in \calL}\ label(\frakb)}
$ \\
\parbox{1cm}{9} \textbf{update} $\calL$, $\calV'$ \\
\parbox{0.5cm}{10} \textbf{for} $\fraka\in\calV\setminus \calL$ \textbf{do}\\
\parbox{1cm}{11} $label(\fraka)=0$ \\
\parbox{0.5cm}{12} \textbf{return} $label(\fraka)$ for $\fraka\in\calV$ \\
\bottomrule
\end{tabularx}

\caption{Labeling of a spreading activation network $\calN(\Delta)$ wrt. a query $\frakq$.}
\label{fig:labeling_of_variables}
\end{figure}

\begin{example}
Figure~\ref{fig:spreading_activation_network} shows
on the left-hand side
the (unlabeled) spreading activation network of $\Delta'$
(Table~\ref{tab:example_knowledge_base}).
The labelings wrt. queries $\frakq_1$ and $\frakq_2$ are shown on the right-hand side. 
For example,
$\Sigma(\frakq_1)=\{a,p\}$ and consequently $label(a)=label(p)=1$.
Next, $b$, $f$, $i$, and $w$
are labeled as they are direct neighbors of at least one of the
atoms $a$, $p$. For instance, $\{a,w\}\in\calE$ and, therefore,
\[
label(w)=
\frac{label(a)}{1+label(a)+label(p)}=\nicefrac{1}{3}.
\]
Atom
$b$ is neighbor of $a$ \emph{and} $p$ and is labeled with
\[
label(b)=\frac{label(a)+label(p)}{1+label(a)+label(p)}=\nicefrac{2}{3}.
\]
\end{example}

Altogether, we are now able to compute $\calA^\Delta_\frakq(\frakr)$ (without usage history).

\begin{example}
\label{ex:activation-based_conditional_inference}
Table \ref{tab:example_knowledge_base_values} shows
$\calA^{\Delta'}_{\frakq_i}$ (cf. also Table \ref{tab:example_knowledge_base})
wrt. the queries $\frakq_1=(p\implies a|\top)$
and $\frakq_2=(\N{f}|c\N{s})$. If a threshold $\theta=2.3$
is used, the conditionals which are selected for
activation-based conditional inference are
\[
\Delta'_1=
({\Delta'})_{\calA_1}^{\theta}=\{\frakr_{1},\frakr_{2},\frakr_{3},\frakr_{6}\},
\]
where $\calA_1=\calA_{\frakq_1}^{\Delta'}$,
and
\[
\Delta'_2=({\Delta'})_{\calA_2}^{\theta}=\{\frakr_{8},\frakr_{9},\frakr_{10},\frakr_{11}\},
\]
where $\calA_2=\calA_{\frakq_2}^{\Delta'}$.
One has 
$[\![\frakq_1]\!]^{\calI^P}_{\Delta'_1}=\textit{yes}$
and
$[\![\frakq_2]\!]^{\calI^P}_{\Delta'_2}=\textit{no}$.
That is, both queries can already 
be decided based on the
reduced belief bases $\Delta'_1$ and $\Delta'_2$ with
activation-based conditional inference.
Note that $\Delta'_1$ and $\Delta'_2$
are smaller than the resp. direct foci according to (standard) focused inference (cf. Example~\ref{ex:focused_inference} for $\frakq_2$). 
\end{example}

In the next section, we make the base-level activation dependent on the history of usage of conditionals and thereby integrate the
concepts of forgetting and remembering into activation-based conditional inference.

\section{Activation-Based Conditional Inference\\ 
and Forgetting and Remembering}

\begin{table}[t]
\begin{tabularx}{\columnwidth}{l|YYY}
\toprule
Conditional &
$\calA^{\Delta'}_{\frakq_2}(\frakr_i)$ &
$\calB^{\Delta'}_{\frakq_1}(\frakr_i)$ &
$\calA_{\frakq_1,\frakq_2}^{\Delta'}(\frakr_i)$  \\
\midrule
$\frakr_1$ & 
$2.27$ & $1.20$ & $\boxed{2.47}$ \\
$\frakr_2$ & 
$2.27$ & $1.20$ & $\boxed{2.47}$ \\
$\frakr_3$ & 
$1.66$ & $1.20$ & $1.86$  \\
$\frakr_4$ & 
$1.70$ & $0.80$ & $1.50$  \\
$\frakr_5$ & 
$1.41$ & $0.80$ & $1.21$  \\
\midrule
$\frakr_6$ & 
$1.60$ & $1.20$ & $1.80$  \\
$\frakr_7$ & 
$1.60$ & $0.40$ & $1.50$  \\
$\frakr_8$ & 
$\boxed{2.43}$ & $0.80$ & $2.23$  \\
$\frakr_9$ & 
$\boxed{2.85}$ & $0.40$ & $\boxed{2.75}$  \\
$\frakr_{10}$ & 
$\boxed{2.94}$ & $0.27$ & $\boxed{2.88}$  \\
\midrule
$\frakr_{11}$ & 
$\boxed{2.58}$ & $0.40$ & $\boxed{2.48}$  \\
$\frakr_{12}$ & 
$1.29$ & $0.80$ & $1.09$  \\
$\frakr_{13}$ & 
$1.12$ & $0.80$ & $0.92$  \\
$\frakr_{14}$ & 
$1.08$ & $0.80$ & $0.88$ \\
$\frakr_{15}$ & 
$1.12$ & $0.80$ & $0.92$ \\
\midrule
$\frakr_{16}$ & 
$1.11$ & $0.80$ & $0.91$   \\
$\frakr_{17}$ & 
$1.07$ & $0.80$ & $0.87$ \\
$\frakr_{18}$ & 
$1.15$ & $0.80$ & $0.95$ \\
$\frakr_{19}$ & 
$1.59$ & $0.40$ & $1.49$  \\
$\frakr_{20}$ & 
$1$ & $0.80$ & $0.80$  \\
\bottomrule
\end{tabularx}
\caption{Activation function
$\calA^{\Delta'}_{\frakq_1,\frakq_2}$
where the base-level activation 
was updated by 
$\phi_{\delta,s}$
with $\delta=0.2$ and
$s^{-1}(1)=(\Delta')^\theta_{\calA^{\Delta'}_{\frakq_1}}$
beforehand.  
$\calA^{\Delta'}_{\frakq_2}$ is recalled
for comparison. Selected conditionals are boxed (threshold $\theta=2.3$).
}
\label{tab:activation-based_inference_and_forgetting}
\end{table}

In ACT-R the base-level activation of a chunk
is not constant but decreases over time and increases
when the chunk is retrieved.
In order to capture this dynamic view on the base-level
activation, we introduce a \emph{forgetting factor}
wrt. a \mbox{selection $s$ by}
\begin{equation}
\phi_{\delta,s}(\frakr)=
\begin{cases}
1+\delta & \IFF s=1 \\
1-\delta & \OTHERWISE
\end{cases}
\end{equation}
with which we update the base-level activation
$\calB^\Delta(\frakr)$ after each inference request.
By doing so, the base-level activation of a conditional is 
decreased when the conditional is not selected for 
answering the query, and it is increased otherwise.
For the updated base-level activation we write
$
\calB_{\delta,s}^\Delta(\frakr)=\calB^\Delta(\frakr)
\cdot \phi_{\delta,s}(\frakr)$.
When applying this update of the base-level activation for every inference request, the usage history of the conditionals is implemented into $\calB^\Delta$ implicitly.

\begin{example}
\label{ex:forgetting}
We compare the activation function $\calA^{\Delta'}_{\frakq_2}$ wrt. query $\frakq_2=(f|c\N{s})$ with the activation function $\calA^{\Delta'}_{\frakq_1,\frakq_2}$ which is obtained by querying $\frakq_1$ first and by updating $\calB^{\Delta'}$ wrt. $s^{-1}(1)=(\Delta')^\theta_{\calA^{\Delta'}_{\frakq_1}}$ and querying $\frakq_2$ afterwards (cf. Table~\ref{tab:activation-based_inference_and_forgetting}, also for parameters,
and Table \ref{tab:example_knowledge_base}).
While in the first case the
conditionals selected for activation-based conditional inference are 
$\{\frakr_8,\frakr_9,\frakr_{10},\frakr_{11}\}$ 
(cf. Example~\ref{ex:activation-based_conditional_inference}), 
in the second case
$
\{\frakr_1,\frakr_2,\frakr_9,\frakr_{10},\frakr_{11}\}$
are selected. In particular, $\frakr_8$ is forgotten because it did not play a role when answering $\frakq_1$.
In both cases, the query $\frakq_2$ is answered with \emph{no}.
\end{example}

The following example shows how remembering is realized within our approach.

\begin{example}
\label{ex:remembering}
When querying $\frakq_1=(p\implies a|\top)$
from $\Delta'$ (Table \ref{tab:example_knowledge_base})
with threshold $\theta=2.3$, the conditional $\frakr_{10}$ is not selected (cf. Table \ref{tab:example_knowledge_base_values}) and consequently its
base-level activation is decreased (cf. Table \ref{tab:activation-based_inference_and_forgetting}).
Afterwards, it has the lowest base-level activation of all conditionals in $\Delta'$. However, it turns out that this conditional is selected and, hence, remembered when asking for $\frakq_2=(f|c\N{s})$ 
afterwards (cf. Example \ref{ex:forgetting}).
\end{example}

Although the base-level activation of a conditional may have been
decreased by the forgetting factor over time to nearly zero, the conditional can still be selected by a selection $s$
if the spreading activation is
high enough to compensate the low base-level activation.

\section{Conclusions and Future Work}
We applied conditional reasoning to ACT-R 
\cite{Anderson_Lebiere_1998,Anderson_2007}
and developed a prototypical model for activation-based conditional inference.
For this, we reformulated
the activation function from ACT-R for conditionals
and selected the conditionals
with the highest degree of activation for focused inference
\cite{Wilhelm_Kern-Isberner_2021}.
With activation-based conditional inference it is possible
to implement several aspects of human reasoning into modern
expert systems such as focusing, 
forgetting, and remembering.

The main challenge for future work is to find for a given query $\frakq$ and a given inference operator $\frakI$ a proper least subset $\Delta'$ of a belief base $\Delta$ such that the query is answered the same wrt. $\Delta'$ as to $\Delta$, i.e. 
$[\![\frakq]\!]^\frakI_{\Delta'}=[\![\frakq]\!]^\frakI_\Delta$, without having to draw the computationally expensive inference $[\![\frakq]\!]^\frakI_\Delta$.

\section*{Acknowledgments}
This work is supported by 
DFG Grant \mbox{KE\,1413/10-1} 
awarded to Gabriele Kern-Isberner 
and DFG Grant \mbox{BE\,1700/9-1} 
awarded to Christoph Beierle 
as part of the priority program 
``Intentional Forgetting in Organizations" (SPP\,1921).

\bibliographystyle{splncs04}
\bibliography{references}

\begin{thebibliography}{10}
\providecommand{\url}[1]{\texttt{#1}}
\providecommand{\urlprefix}{URL }
\providecommand{\doi}[1]{https://doi.org/#1}

\bibitem{Adams_1975}
Adams, E.W.: The Logic of Conditionals. Springer (1975)

\bibitem{Anderson_1980}
Anderson, J.R.: Human Associative Memory: A Brief Edition. Lawrence Erlbaum
  Assoc. (1980)

\bibitem{Anderson_1983}
Anderson, J.R.: A spreading activation theory of memory. Journal of Verbal
  Learning and Verbal Behavior  \textbf{22},  261--295 (1983)

\bibitem{Anderson_2007}
Anderson, J.R.: How can the human mind occur in the physical universe? Oxford
  University Press (2007)

\bibitem{Anderson_Lebiere_1998}
Anderson, J.R., Lebiere, C.: The atomic components of thought. Psychology Press
  (1998)

\bibitem{Goldszmidt_Pearl_1991}
Goldszmidt, M., Pearl, J.: On the consistency of defeasible databases. Artif.
  Intell.  \textbf{52}(2),  121--149 (1991)

\bibitem{Jackson_1999}
Jackson, P.: Introduction to Expert Systems. Addison Wesley, 3rd edn. (1999)

\bibitem{Kern-Isberner_Beierle_Brewka_2020}
Kern{-}Isberner, G., Beierle, C., Brewka, G.: Syntax splitting = relevance +
  independence: New postulates for nonmonotonic reasoning from conditional
  belief bases. In: Calvanese, D., Erdem, E., Thielscher, M. (eds.) Proceedings
  of the 17th International Conference on Principles of Knowledge
  Representation and Reasoning, {KR} 2020. pp. 560--571 (2020)

\bibitem{Klahr_Langley_Neches_1987}
Klahr, D., Langley, P., Neches, R. (eds.): Production System Models of Learning
  and Development. MIT Press (1987)

\bibitem{Kraus_Lehmann_Magidor_1990}
Kraus, S., Lehmann, D., Magidor, M.: Nonmonotonic reasoning, preferential
  models and cumulative logics. Artif. Intell.  \textbf{44}(1-2),  167--207
  (1990)

\bibitem{Newell_1990}
Newell, A.: Unified Theories of Cognition. Harvard University Press (1990)

\bibitem{Pearl_1990}
Pearl, J.: System {Z:} {A} natural ordering of defaults with tractable
  applications to nonmonotonic reasoning. In: Parikh, R. (ed.) Proceedings of
  the 3rd Conference on Theoretical Aspects of Reasoning about Knowledge. pp.
  121--135. Morgan Kaufmann (1990)

\bibitem{Rajendra_Sajja_2009}
Rajendra, A., Sajja, P.: Knowledge-Based Systems. Jones and Bartlett Learning
  (2009)

\bibitem{Ryle_2000}
Ryle, G.: The Concept of Mind. University of Chicago Press, {N}ew edn. (2000)

\bibitem{Spohn_2012}
Spohn, W.: The Laws of Belief: Ranking Theory and Its Philosophical
  Applications. Oxford University Press (2012)

\bibitem{Wilhelm_Kern-Isberner_2021}
Wilhelm, M., Kern{-}Isberner, G.: Focused inference and {S}ystem {P}. In:
  Thirty-Fifth {AAAI} Conference on Artificial Intelligence, {AAAI} 2021. pp.
  6522--6529. {AAAI} Press (2021)

\end{thebibliography}
\end{document}